\documentclass[sigconf,natbib=false]{acmart}

\AtBeginDocument{%
  }

\setcopyright{acmcopyright}
\copyrightyear{2022}
\acmYear{2022}
\acmDOI{XXXXXXX.XXXXXXX}

\usepackage{amsmath}
\usepackage{amssymb}
\usepackage{verbatim}
\usepackage{multirow}
\usepackage{amsthm,amsmath,amssymb}
\usepackage{mathrsfs}
\usepackage{subfigure}

\RequirePackage[
  datamodel=acmdatamodel,
  style=acmnumeric,
  ]{biblatex}

\begin{document}

\title{Detecting and Recovering Adversarial Examples from Extracting  Non-robust and Highly Predictive Adversarial Perturbations}

\author{Mingyu Dong}
\orcid{0000-0002-4049-5883}
\affiliation{%
  \institution{Ningbo University}
  \city{Ningbo}
  \state{Zhejiang}
  \country{CN}
  \postcode{315211}
}
\email{1253173217@qq.com}

\author{Jiahao Chen}
\affiliation{%
  \institution{Ningbo University}
  \city{Ningbo}
  \state{Zhejiang}
  \country{CN}
  \postcode{315211}
}
\email{196003641@nbu.edu.cn}

\author{Diqun Yan}
\authornote{Both authors contributed equally to this research.}
\affiliation{%
  \institution{Ningbo University}
  \city{Ningbo}
  \state{Zhejiang}
  \country{CN}
  \postcode{315211}
}
\email{yandiqun@nbu.edu.cn}

\author{Jinxing Gao}
\affiliation{%
  \institution{Ningbo University}
  \city{Ningbo}
  \state{Zhejiang}
  \country{CN}
  \postcode{315211}
}
\email{2011082290@nbu.edu.cn}

\author{Li Dong}
\affiliation{%
  \institution{Ningbo University}
  \city{Ningbo}
  \state{Zhejiang}
  \country{CN}
  \postcode{315211}
}
\email{dongli@nbu.edu.cn}

\author{Rangding Wang}
\affiliation{%
  \institution{Ningbo University}
  \city{Ningbo}
  \state{Zhejiang}
  \country{CN}
  \postcode{315211}
}
\email{wangrangding@nbu.edu.cn}

\renewcommand{\shortauthors}{Trovato et al.}

\begin{abstract}
Deep neural networks (DNNs) have been shown to be vulnerable against adversarial examples (AEs) which are maliciously designed to fool target models. The normal examples (NEs) added with imperceptible adversarial perturbation, can be a security threat to DNNs. Although the existing AEs detection methods have achieved a high accuracy, they failed to exploit the information of the AEs detected. Thus, based on high-dimension perturbation extraction, we propose a model-free AEs detection method, the whole process of which is free from querying the victim model. Research shows that DNNs are sensitive to the high-dimension features. The adversarial perturbation hiding in the adversarial example belongs to the high-dimension feature which is highly predictive and non-robust. DNNs learn more details from high-dimension data than others. In our method, the perturbation extractor can extract the adversarial perturbation from AEs as high-dimension feature, then the trained AEs discriminator determines whether the input is an AE. Experimental results show that the proposed method can not only detect the adversarial examples with high accuracy, but also detect the specific category of the AEs. Meanwhile, the extracted perturbation can be used to recover the AEs to NEs. 

\end{abstract}



\maketitle

\section{Introduction}
\label{sec1}
Deep Neural Networks (DNNs) have achieved state-of-the-art performance in a wide range of real-world applications, such as speech recognition, image classification, object detection, etc\cite{2018A}.
However, it has shown by recent studies that DNNs can be fooled easily by adversarial examples, which are crafted by maliciously adding designed perturbations. Despite the imperceptibility of adversarial examples (AEs), these intentionally-perturbed inputs can induce the network to make incorrect predictions with high confidence \cite{2019Adversarial}. This undesirable phenomenon of deep networks has raised much concern about the security of DNNs in real-world applications, such as self-driving cars and identity recognition. The countermeasures for adversarial examples have also been proposed, which can be roughly categorized into following types: adversarial detection, robust models and data preprocessing. In this paper, we intend to reveal the properties of adversarial perturbations, and exploit them for adversarial detection.

Even though there has been many robust classification strategies\cite{gou2021knowledge}, most of them are still not effective when the AEs are generated by some advanced attacks like \cite{carlini2017towards}. Therefore, defending against adversarial attack is the top priority. Comparing with the other two methods, adversarial detection cost less without sacrificing the accuracy of the models. In many practical applications, the crafted samples will be distinguished by detector and then can guide a better defense strategy.

\begin{figure}[h]
  \centering
  \includegraphics[width=1\linewidth]{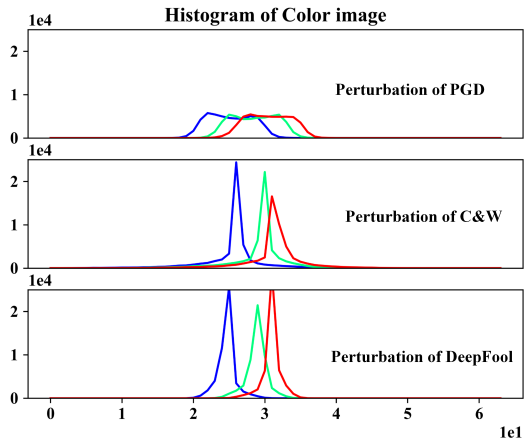}
  \caption{Histogram of color on three different adversarial perturbations.}
  \label{color}
\end{figure}

In recent years, research on AEs has caused widespread concern. AEs are geometrically close to normal examples, but they are maliciously designed to cross the decision boundary with minimized perturbation magnitude, as proposed by Szegedy in 2013 \cite{szegedy2013intriguing}. According to the dependence of the generation process on victim models, the threat models can be divided into three parts, white-box, grey-box, and black-box attack. In white-box attacks, the parameters and structure of the model are known by attackers. The gradient-based attack is the classical method that directly adds the sign of gradients to the normal examples \cite{goodfellow2014explaining} and iterative generation process can help improve the attack success rate (ASR) but get more obvious adversarial perturbation \cite{madry2017towards}. The design of optimize-based methods is to optimize the generated adversarial examples to have fewer perturbations but keep the high attack rate \cite{carlini2017towards, moosavi2016deepfool}. Under black-box settings, the parameters and structure of the target model are not accessible such as transferable adversarial attack, and it is more practical in the real world.

\begin{figure*}[t]
\centering 
\includegraphics[width=0.9\textwidth]{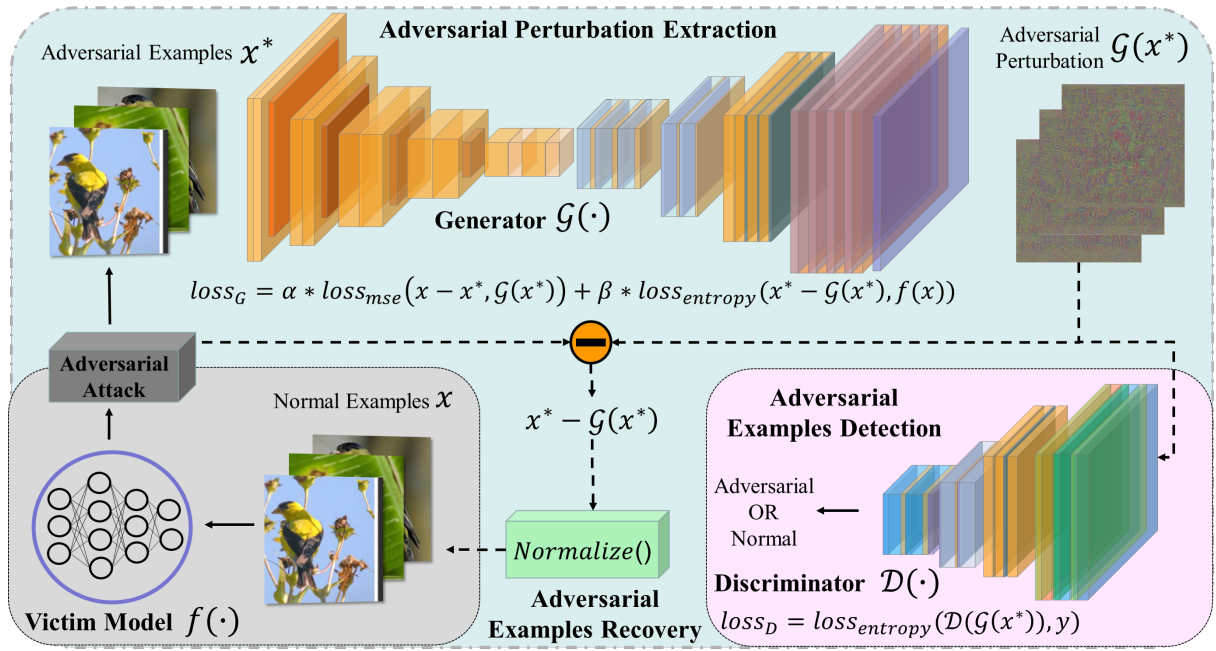} %
\caption{Overview of the adversarial perturbation extracting and perturbation discriminating. The adversarial examples  are generated from trained victim models. The adversarial examples are put into the APE to extract the adversarial perturbation. In adversarial examples detection, the extracted adversarial perturbations are discriminated by the discriminator. In adversarial examples recovery, the extracted perturbations are subtracted from adversarial examples, after normalizing, the adversarial examples are recovered. } 
\label{Fig.main2} 
\end{figure*}

The existence of adversarial examples seriously affects the robustness of the DNNs. To solve this problem, previous work has proposed a variety of explanations. Meng and Chen constructed a denoiser model to transform adversarial examples into normal examples, and the reconstruction error was calculated as a score to judge whether the given sample is an adversarial example or not \cite{meng2017magnet}. Lewis Smith projected the uncertainty of the adversarial example onto lower-dimensional spaces, and improved the quality of uncertainty estimates using probabilistic model ensembles \cite{smith2018understanding}. Freitas extracted the object-based features from the input sample, and these low-level features in the adversarial example contains the information of the target category \cite{freitas2020unmask}, so the comparison of the low-level features between the original category and target category can help to detect whether it is an adversarial example. Andrew proposed that adversarial examples are not bugs, but features and validated their hypothesis in \cite{ilyas2019adversarial} and they revealed that adversarial vulnerability did not stem from using a specific model class or a specific training method. The noise added into the normal example is what we called adversarial perturbation and can be viewed as a special artifact in the input sample. The classifiers can be more robust if they are learned by highly predictive features. However, the guide of non-robust features will have the models overlearned. These high-dimension artifacts are sensitive for DNNs and tend to make the neural network misjudge \cite{yuan2019adversarial}, while these kind of features are almost imperceptible to humans.

Previous work has laid a solid foundation for us, inspired by the previous studies above. We find that most of the existing methods of AEs detection need to query the victim model to get some deep features or logits output of the classification results. To get rid of the dependence on the victim models, we propose a model-free method to detect adversarial example from the angle of the adversarial perturbation itself even the protected model is not accessible. The high-dimension information in the adversarial examples is fully exploited to learn the feature of AEs.

In this paper, we present an adversarial perturbation extractor to extract the high-dimension feature of the AEs. The detector acts as a discriminator to distinguish the extracted perturbation and help the extractor to converge quickly and extract precisely. The fundamental idea is to simulate the adversarial perturbation hiding in examples by a reconstruction network and then find the differences between the normal and the adversarial examples by a discriminator. Experimental results have shown that the perturbations extracted from the NEs and AEs are obviously different, which can be easily distinguished by the discriminator.
The recovered samples can be obtained by subtracting extracted adversarial perturbations from AEs.

Before experiments, the statistical histogram of color on different adversarial perturbations is shown in Figure \ref{color}. According to the statistical results we have, the deviation on adversarial perturbations between the same attack methods is less than that in different attack methods, even the adversarial examples are generated from different victim models. The adversarial perturbations in the different attack methods are different visually as shown in Figure \ref{image_per}. Thus the adversarial perturbation can be the gist of detection. Based on this observation, a multi-classification discriminator is proposed to identify the attack methods operated on NEs instead of a binary detector in previous works. The specific classification of adversarial examples can help the victim model effectively carry out specific defense strategies against the attack of this type of adversarial examples.

The contributions of our work in this paper can be concluded as follows:
\begin{itemize}

\item
    We propose model-free adversarial example detection for the detection of AEs to protect DNNs under both white-box and black-box settings and it is more applicable in the real world for it is model-free. The excellent performance of the multi-classification discriminator have also shown the superiority of our work.
    
\item
    Our extensive experiments on adversarial perturbation extraction reveal the properties of AEs and validate the hypothesis that adversarial perturbations are the features which are non-robust but highly predictive. From the perspective of interpretability, they are the intrinsic properties of models.

\item
    Considering the original value of NEs, the perturbations extracted can help to have the samples, which are crafted, recovered. The recovery experiments show that we can take advantage of the AEs again instead of just ignoring them in the previous studies.

\end{itemize}

The rest of this paper is organized as follows: Related works about AEs detection will be introduced in Section 2. Our proposed method will be illustrated in Section 3. In Section 4, we will introduce the experimental setup and explain the results of the experiments. In Section 5, we summarise our work of this paper and propose the directions for future improvement.

\section{Adversarial example detection}

This section provides a brief introduction to the existing AEs detection method. According to the dependency on victim models, the detection task on the adversarial examples can be described as a binary classification on the whole. A variety of methods have been proposed, which can be divided into unsupervised and supervised methods.

\subsection{Supervised method}
Under supervised seetings, the discriminator is trained by adversarial examples generated from trained victim models. Regarding the adversarial example as a new category to retrain the model, Kathrin at el. put the adversarial example into the model directly to detect the abnormal examples\cite{grosse2017statistical,hosseini2017blocking}.   Aigrain used the logits of original examples and adversarial examples from trained models to learn a binary classifier to distinguish them \cite{aigrain2019detecting}. The artifacts in different samples can behave at different densities: those in adversarial examples subspaces are lower than that of the original examples. Based on this theory, Feinman proposed Kernel Density estimation for each class in training data and then trained a binary classifier\cite{feinman2017detecting}. Another work extracted the deep features by putting original examples and adversarial examples into a trained model, in which the difference of theses deep features between AEs and NEs was learned \cite{sperl2020dla}. \cite{abusnaina2021adversarial} presented the first graph-based adversarial detection method in which a latent neighborhood graph and a graph attention discriminator around input samples was constructed to determine whether the input samples are AEs. The supervised methods need to query the victim model more or less, many methods concentrate on the deep feature or the related classification information from the output of the victim model.  

\subsection{Unsupervised method}
Under unsupervised settings, the detector is only accessible to original examples. In this scene, the usual method is to enlarge the difference between adversarial examples and original examples. MagNet was proposed and was trained by normal examples only, in which the adversarial examples can be detected according to the difference of the reconstruction error with a reconstruct network \cite{meng2017magnet}. Carrara et al. came out with a $k$-NN classifier to score the category of the output from a trained victim model and detected adversarial examples by a specified threshold \cite{carrara2017detecting}. Reverse cross-entropy was used to retrain the victim classifier for learning the deep feature of normal and adversarial examples. The adversarial ones can be detected by a trained threshold \cite{pang2018towards}. Liang at el. proposed a feature squeezing approach: they used the image entropy as a metric to implement the adaptive noise reduction, and the detector would judge the input as a adversarial one if the output result is different from the original one \cite{liang2018detecting}. Color bit-depth reduction and local smoothing also belong to feature  squeeze methods \cite{xu2017feature}. The unsupervised methods usually lack the training process, through the difference in threshold compared with the normal examples.  

\section{Methology}

The proposed detection mechanism concentrates on extracting the high-dimension features of the adversarial examples from which the adversarial perturbations are extracted. Since DNNs are sensitive to high-dimension features, the discriminator will learn the differences of different attack methods projected on high dimension. The perturbation in normal example and adversarial example will be extracted by a U-Net structure to exploit the perturbation textures of AEs and NEs with the guidance of discriminator, then the extracted perturbation will be forwarded into the discriminator. The overview of the proposed detection mechanism is shown in Figure 1.

\subsection{Perturbation Extration}
An adversarial example can be simply separated into a normal example and an adversarial perturbation theoretically. However, to distinguish the adversarial examples directly is hard. Instead, we can tell the difference from the perceptive of adversarial perturbations easily, so the design of U-Net is to extract the perturbation from the samples. The structure of U-Net has been widely applied in many reconstruction tasks. For example, Chen proposed a method that used U-Net to extract OSN\cite{wu2022robust}. The structure of U-Net in our method, which can be seen in Figure 1, is common and easy to reproduce. 
Five downsample encoder blocks are used, following five upsample decoder blocks in U-Net. In every encoder block,  there are two two-dimensional convolutional layers, two ReLU \cite{nair2010rectified} activation layers, and two two-dimensional BatchNomalization. After one downsample block, the MaxPooling layer will reduce the shape of the data. In the encoding process, the 5 blocks will make the size of the input data get flatter, and the high-dimension feature of the adversarial example will be squeezed into a 512-channel and small matrix whose length and width are 28-dimension. In each decoder block, 1 two-dimensional convolutional layer, 1 ReLU activation layer, and 1 two-dimensional BatchNomalization are contained. In the decoding process, two data will be concatenated by Skip Connection to enlarge the features. At last, the shape of the output will be as same as input.

For each training iteration, we first sample two pristine 3-channel (RGB) color images which contains input examples $X$ and ground truth of perturbations $Y=X-x$, where $x$ are the normal examples. The $Y$ are the target of the U-Net. To reconstruct the adversarial perturbation, the target of U-Net is set as the ground truth perturbation to assure the extracted perturbation is close to the ground truth. Additionally, to ensure that the quality of recovered images, the error between adversarial perturbation and the ground truth should also be restricted, so the Mean Square Error (MSE) is considered:
\begin{equation}
    loss_{MSE}=\frac{1}{N}\sum_{i=1}^{N}MSE(\mathcal{G} (X_{i}), Y_{i})
\end{equation}
where $N$ is the number of input examples in a batch; $X_{i}$ is the input example; $\mathcal{G}()$ represents the U-Net. For NEs $x$, the MSE loss guarantees that $\mathcal{G}(x)$ is close to 0. In the meantime, for crafted examples $x^{*}$, the MSE loss guarantees that the $x^{*}-\mathcal{G}(x^{*})$ is close to NEs $x$. For the consideration of the victim classifier, the output of the adversarial examples subtracts the generated adversarial perturbation should also be correctly classified $f(x^{*}-\mathcal{G}(x^{*}))=y$ to guarantee the quality of the recovered image. To ensure that the extracted perturbations are non-robust but highly predictive, which means that the victim model is vulnerable to these perturbations, the cross-entropy loss is used:
\begin{equation}
    loss_{CE}=\frac{1}{N}\sum_{i=1}^{N}\sum_{j=1}^{C}-[y_{ij}\ln f(x_{ij}-\mathcal{G}(x_{ij}))]
\end{equation}
where $y$ is the output of the victim models given NEs, $C$ is quantity of the category. For NEs $x$, the cross-entropy loss guarantees that the outputs of $x-\mathcal{G}(x)$ and $x$ are consistent. In the meantime, for crafted examples $x^{*}$, the cross-entropy loss guarantees that the recovered examples $x^{*}-\mathcal{G}(x^{*})$ can be classified correctly. The loss of cross-entropy will not update the parameters of victim classifier. The total loss function, combined with the corresponding coefficients, can be described as follows:
\begin{equation}
    loss=\alpha*loss_{MSE}+\beta*loss_{CE}
\end{equation}

The weighted sum of these two loss functions will help the U-Net to extract the adversarial perturbation as similar to the ground truth perturbation.

\subsection{Perturbation Discriminator}
\begin{figure}
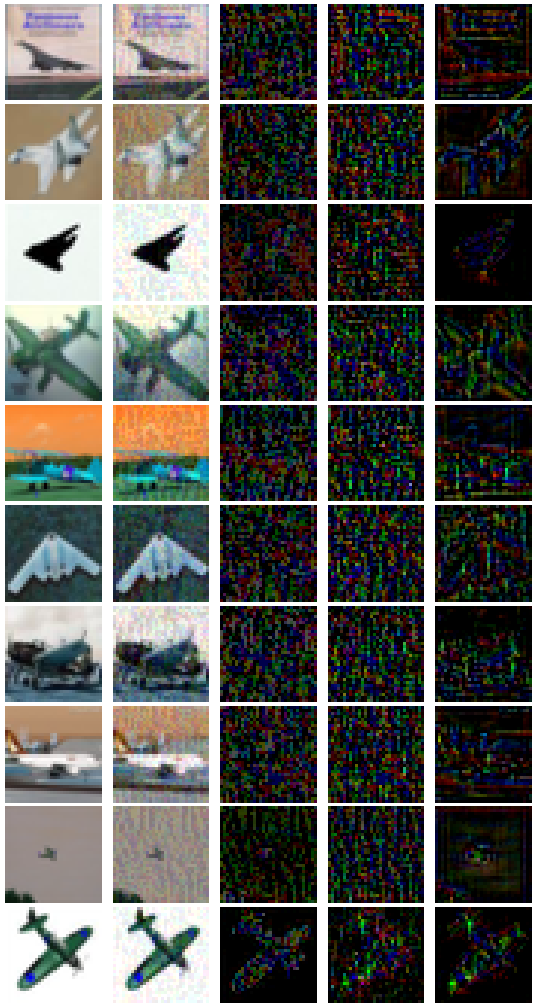

	\centering
	\subfigure{
		\begin{minipage}[t]{0.15\linewidth}
			\centering
			\includegraphics[width=0.5in]{3.pdf}\\
			\vspace{0.03cm}
			\includegraphics[width=0.5in]{10.pdf}\\
			\vspace{0.03cm}
			\includegraphics[width=0.5in]{21.pdf}\\
			\vspace{0.03cm}
			\includegraphics[width=0.5in]{27.pdf}\\
			\vspace{0.03cm}
			\includegraphics[width=0.5in]{44.pdf}\\
			\vspace{0.03cm}
			\includegraphics[width=0.5in]{52.pdf}\\
			\vspace{0.03cm}
			\includegraphics[width=0.5in]{74.pdf}\\
			\vspace{0.03cm}
			\includegraphics[width=0.5in]{90.pdf}\\
			\vspace{0.03cm}
			\includegraphics[width=0.5in]{97.pdf}\\
			\vspace{0.03cm}
			\includegraphics[width=0.5in]{98.pdf}\\
			\vspace{0.03cm}
		\end{minipage}%
	}
	\subfigure{
		\begin{minipage}[t]{0.15\linewidth}
			\centering
			\includegraphics[width=0.5in]{3_pgd_eps0.1_0_8.pdf}\\
			\vspace{0.03cm}
			\includegraphics[width=0.5in]{10_pgd_eps0.1_0_4.pdf}\\
			\vspace{0.03cm}
			\includegraphics[width=0.5in]{21_pgd_eps0.1_0_2.pdf}\\
			\vspace{0.03cm}
			\includegraphics[width=0.5in]{27_pgd_eps0.1_0_4.pdf}\\
			\vspace{0.03cm}
			\includegraphics[width=0.5in]{44_pgd_eps0.1_0_4.pdf}\\
			\vspace{0.03cm}
			\includegraphics[width=0.5in]{52_pgd_eps0.1_0_7.pdf}\\
			\vspace{0.03cm}
			\includegraphics[width=0.5in]{74_pgd_eps0.1_0_6.pdf}\\
			\vspace{0.03cm}
			\includegraphics[width=0.5in]{90_pgd_eps0.1_0_8.pdf}\\
			\vspace{0.03cm}
			\includegraphics[width=0.5in]{97_pgd_eps0.1_0_3.pdf}\\
			\vspace{0.03cm}
			\includegraphics[width=0.5in]{98_pgd_eps0.1_0_7.pdf}\\
			\vspace{0.03cm}
		\end{minipage}%
	}
	\subfigure{
		\begin{minipage}[t]{0.15\linewidth}
			\centering
			\includegraphics[width=0.5in]{3_ground_per.pdf}\\
			\vspace{0.03cm}
			\includegraphics[width=0.5in]{10_ground_per.pdf}\\
			\vspace{0.03cm}
			\includegraphics[width=0.5in]{21_ground_per.pdf}\\
			\vspace{0.03cm}
			\includegraphics[width=0.5in]{27_ground_per.pdf}\\
			\vspace{0.03cm}
			\includegraphics[width=0.5in]{44_ground_per.pdf}\\
			\vspace{0.03cm}
			\includegraphics[width=0.5in]{52_ground_per.pdf}\\
			\vspace{0.03cm}
			\includegraphics[width=0.5in]{74_ground_per.pdf}\\
			\vspace{0.03cm}
			\includegraphics[width=0.5in]{90_ground_per.pdf}\\
			\vspace{0.03cm}
			\includegraphics[width=0.5in]{97_ground_per.pdf}\\
			\vspace{0.03cm}
			\includegraphics[width=0.5in]{98_ground_per.pdf}\\
			\vspace{0.03cm}
		\end{minipage}%
	}
	\subfigure{
		\begin{minipage}[t]{0.15\linewidth}
			\centering
			\includegraphics[width=0.5in]{3_adv_per.pdf}\\
			\vspace{0.03cm}
			\includegraphics[width=0.5in]{10_adv_per.pdf}\\
			\vspace{0.03cm}
			\includegraphics[width=0.5in]{21_adv_per.pdf}\\
			\vspace{0.03cm}
			\includegraphics[width=0.5in]{27_adv_per.pdf}\\
			\vspace{0.03cm}
			\includegraphics[width=0.5in]{44_adv_per.pdf}\\
			\vspace{0.03cm}
			\includegraphics[width=0.5in]{52_adv_per.pdf}\\
			\vspace{0.03cm}
			\includegraphics[width=0.5in]{74_adv_per.pdf}\\
			\vspace{0.03cm}
			\includegraphics[width=0.5in]{90_adv_per.pdf}\\
			\vspace{0.03cm}
			\includegraphics[width=0.5in]{97_adv_per.pdf}\\
			\vspace{0.03cm}
			\includegraphics[width=0.5in]{98_adv_per.pdf}\\
			\vspace{0.03cm}
		\end{minipage}%
	}
	\subfigure{
		\begin{minipage}[t]{0.15\linewidth}
			\centering
			\includegraphics[width=0.5in]{3_ori_per.pdf}\\
			\vspace{0.03cm}
			\includegraphics[width=0.5in]{10_ori_per.pdf}\\
			\vspace{0.03cm}
			\includegraphics[width=0.5in]{21_ori_per.pdf}\\
			\vspace{0.03cm}
			\includegraphics[width=0.5in]{27_ori_per.pdf}\\
			\vspace{0.03cm}
			\includegraphics[width=0.5in]{44_ori_per.pdf}\\
			\vspace{0.03cm}
			\includegraphics[width=0.5in]{52_ori_per.pdf}\\
			\vspace{0.03cm}
			\includegraphics[width=0.5in]{74_ori_per.pdf}\\
			\vspace{0.03cm}
			\includegraphics[width=0.5in]{90_ori_per.pdf}\\
			\vspace{0.03cm}
			\includegraphics[width=0.5in]{97_ori_per.pdf}\\
			\vspace{0.03cm}
			\includegraphics[width=0.5in]{98_ori_per.pdf}\\
			\vspace{0.03cm}
		\end{minipage}%
	}
	\centering
	\caption{Five columns are normal images, adversarial examples, ground truth adversarial perturbation, extracted perturbation from adversarial examples and extracted perturbation from original examples, respectively. Images are all chosen from CIFAR-10.}
	\vspace{-0.2cm}
	\label{cifar_per}
\end{figure}

The extracted perturbation from normal and adversarial examples can be seen in Figure \ref{cifar_per}. From the third colum ,the fourth colum  and the fifth colum of the figure, we can see that the perturbations extracted from the adversarial examples are close to the ground truth perturbation, but the perturbations extracted from the normal examples are different from the extracted adversarial examples. DNNs are sensitive to high-dimension features \cite{yuan2019adversarial}, adversarial perturbation belongs to the high-dimension features. The normal example added with high-dimension adversarial perturbation can confuse the classifier. Inspired by this hypothetical, the design of the discriminator to detect the difference in extracted perturbation may work, and the sensitivity of DNNs may help to learn more specific information than using adversarial examples directly. The obvious difference between normal and adversarial perturbation can be enlarged in the high dimension.

After extracting perturbations from adversarial examples and normal examples, the extracted perturbations will be put into the discriminator as images. To meet the requirements of a lightweight model, we choose a simple binary discriminator $\mathcal{D}$ to detect the extracted perturbations. The choice of loss function is cross-entropy, it can be seen as follows:
\begin{equation}
    loss=\frac{1}{N}\sum_{i=1}^{N}-[z_{i}\ln \mathcal{D}(x-\mathcal{G}(x))+(1-z_{i})*\ln(1-\mathcal{D}(x-\mathcal{G}(x)))]
\end{equation}
where $z$ denotes the categories of the input samples including 0 and 1. The label of the adversarial example is set as 1, and the normal example is 0. For the detection of different adversarial attacks, this binary detector could be extended to a multi-classification detector.

\begin{table}[htbp]
\caption{Performance of trained victim image classifiers on different data set and models are shown in the right. The test accuracy of three adversarial attack methods are shown in the right.}
\centering
\begin{tabular}{llcccc}
\toprule
\multirow{2}{*}{DataSet} & \multirow{2}{*}{Model Name} & \multicolumn{4}{c}{Test Accuracy(\%)}\\ \cline{3-6}
          &         & original & PGD & C\&W & DeepFool \\ 
\midrule
\multirow{3}{*}{CIFAR-10}  & MovileNetV2  & 67.73 & 21.73 & 10.37 & 20.16  \\
                           & ResNet50     & 66.19 & 23.54 & 9.79 & 20.16    \\
                           & VGG19        & 79.85 & 15.27 & 9.79 & 20.15     \\
\multirow{3}{*}{ImageNet}  & MovileNetV2  & 65.73 & 4.71 & 3.33 & 11.54       \\
                           & ResNet50     & 62.97 & 6.28 & 3.40 & 12.41        \\ 
                           & VGG19        & 60.70 & 6.22 & 3.42 & 13.21         \\
\bottomrule
\end{tabular}
\label{victim_train}
\end{table}

\section{Experiment}

In this section, we conducted several experiments to test the detection capacity of proposed method. Some other usages of the extracted perturbations are also considered. In summary, we design the following experiments: 

\noindent
\textbf{Detection Accuracy}: We train the U-Net and discriminator only by one attack method, and then test it to detect unseen adversarial examples generated by different models and attack methods. 

\noindent
\textbf{Comarision with state-of-the-art}: The U-Net and discriminator are trained by several attack methods. We compare the detection rate of proposed methods with some state-of-the-art detection methods, include testing on other unseen attack methods. 

\noindent
\textbf{Adversarial examples recover}: To verify the extracted adversarial perturbation can represent the high-dimension feature of the adversarial examples, we subtract the extracted adversarial perturbations from the adversarial examples to recover the adversarial examples.

\noindent
\textbf{Adversarial examples classification}: We expend the usage of the extracted perturbation. To detect what kind of attack method is, we design the adversarial perturbations classification experiments.

\noindent
\textbf{Robustness test}: To test the robustness of adversarial examples detection methods, we add some random noise to NEs of the test set to simulate the distortion in the real world.

\begin{table*}[t]
\caption{Adversarial examples detected rate between different models. The training data of proposed method is generated from one model and single attack method. The $DL$ in table is DeepFool.}
\centering
\begin{tabular}{llccccccccc}
\toprule
    &    & \multicolumn{3}{c}{MobileNetV2} & \multicolumn{3}{c}{ResNet50} & \multicolumn{3}{c}{VGG19}\\ \cline{3-11} 
    & & PGD(\%) & CW(\%) & DL(\%) & PGD(\%) & CW(\%) & DL(\%) & PGD(\%) & CW(\%) & DL(\%)  \\ \midrule
    
\multicolumn{1}{l}{\multirow{3}{*}{MobileNetV2}} & PGD & 93.48 & 93.40 & 55.03 & 93.58 & 93.35 & 54.88 & 93.53 & 92.50 & 54.83     \\
\multicolumn{1}{c}{}& CW  & 50.00 & 99.98 & 50.00 & 50.00 & \textbf{100.00} & 50.00 & 50.00 & 99.98 & 50.00     \\ 
\multicolumn{1}{c}{}& DF  & 50.03 & 49.80 & 89.30 & 49.75 & 49.83 & 89.88 & 50.00 & 49.78 & 89.35  \\ 

\multicolumn{1}{l}{\multirow{3}{*}{ResNet50}} & PGD & \textbf{100.00} & 99.35 & 52.98 & 99.88 & 99.28 & 53.53 & \textbf{100.00} & 99.3 & 53.65  \\
\multicolumn{1}{c}{} & CW  & 50.01 & \textbf{100.00} & 50.00 & 50.00 & \textbf{100.00} & 50.00 & 50.01 & \textbf{100.00} & 50.00    \\ 
\multicolumn{1}{c}{}& DF  & 50.05 & 50.15 & 89.43 & 50.03 & 50.28 & 89.85 & 50.00 & 50.20 & 89.68  \\ 

\multicolumn{1}{l}{\multirow{3}{*}{VGG19}} & PGD & \textbf{100.00} & 99.78 & 51.13 & \textbf{100.00} & 99.53 & 51.33 & 99.98 & 99.83 & 51.58 \\
\multicolumn{1}{c}{} & CW & 50.00 & \textbf{100.00} & 50.00 & 50.00 & \textbf{100.00} & 50.00 & 50.00 & 99.96 & 50.00 \\ 
\multicolumn{1}{c}{}& DF  & 50.03 & 49.80 & 89.30 & 49.75 & 49.83 & 89.88 & 50.00 & 49.78 & 89.35  \\ 
\bottomrule
\end{tabular}
\label{detct_acc}
\end{table*}

\subsection{Experiment setup}
Our experiments are conducted on the CIFAR-10 \cite{krizhevsky2009learning} and ImageNet \cite{deng2009imagenet} datasets. MobileNetV2 \cite{2018Inverted}, ResNet50 \cite{2014Very}, and VGG19 \cite{2016Deep} are chosen to be trained as the victim image classifiers, the accuracy on test sets can be seen in Table \ref{victim_train}. FGSM, PGD, C\&W, and DeepFool are chosen as attack methods which are generated from the trained models above. AEs can reduce the accuracy of victim classifiers, the results are also shown in Table \ref{victim_train}. The initial learning rate of U-Net and victim image classifier is set to 0.0001, and the epoch of U-Net and adversarial discriminator is 300 and 100, respectively, the values of $\alpha$ and $\beta$ in loss function of the U-Net are all 1. Adam Optimizer is chosen to optimize the training process.

\subsection{Detection accuracy}

In this section, we test the detection rate of the proposed method. Firstly, the U-Net is trained by AEs which are generated from one single victim classifier model. Then the adversarial discriminator is trained by adversarial perturbations and the corresponding normal perturbation which are generated by U-Net. The test set is composed of 50\% adversarial examples which are generated from same model and 50\% normal examples. These experiments are conducted in the CIFAR-10 date set.

We also consider the generalization of the method. According to our hypothetical, the same attack generated from different models should have a similar adversarial perturbation, so we prepare enough adversarial examples generated from different models and using different attack methods. We train the U-Net and discriminator only by a single attack method which is generated from one victim model. The test set is generated by other trained victim models and attack methods, where the adversarial examples and original examples are accounts for 50\%, respectively.

The generalization to other unseen adversarial examples is also considered. The training process is as same as above. And the test sets are in the same configuration as above, but the AEs are generated from other models or using different attack methods.

The results of the experiments are shown in Table \ref{detct_acc}, it is the detection accuracy of AEs. In every row of the table, the U-Net and the discriminator are trained by the data generated from one victim model and in one attack method. In Every column of the table, the test set is generated from different victim models and different attack methods. We can see that the same attack methods can be detected with a high accuracy, even the AEs are generated by different victim models. It means that different AEs which generated from different victim models still have similar adversarial perturbations in the same attack method. But the proposed method trained by PGD can detect the unseen adversarial examples, but that trained by C\&W cannot distinguish other methods except for itself. According to the extracted adversarial perturbations by U-Net, the distribution of PGD perturbation is less complex than that of C\&W, the classifier trained by C\&W will detect the PGD as a normal one, thus the detection results are all close to  50\%. 

\begin{table*}[htbp]
\caption{Experimental results of Comparison, three SOTA and classic methods are chosen to compare with the proposed method. The detection accuracy and f1Score are chosen to measure effect of all methods. These experiments are conduct in ImageNet dataset.}
\centering
\begin{tabular}{lccccccc}
\toprule
         & methods & \multicolumn{2}{c}{MobileNetV2} & \multicolumn{2}{c}{ResNet50} & \multicolumn{2}{c}{VGG19} \\ \cline{3-8} 
  & & ACC(\%) & f1Score(\%) & ACC(\%) & f1Score(\%) & ACC(\%) & f1Score(\%)   \\ 
\midrule
\multirow{4}{*}{MobileNetV2} & MagNet         & 58.20 & 48.52  & 57.73 & 48.24 & 57.84 & 48.45 \\
                             & DLA            & 90.74 & 90.35 & 77.60 & 72.75 & 77.51 & 72.61      \\
                             & PACA           & 99.98 & 99.98 & 99.97 & 99.97 & 99.99 & 99.99      \\
                             & proposed       & 99.97 & 99.97 & 99.97 & 99.97 & 99.97 & 99.97         \\ 
                             \hline
\multirow{4}{*}{ResNet50}    & MagNet         & 58.20 & 48.52  & 57.73 & 48.24 & 57.84 & 48.45 \\
                             & DLA            & 81.02 & 76.99 & 92.97 & 92.58 & 92.57 & 92.12         \\
                             & PACA           & 99.98 & 99.98 & 99.97 & 99.97 & 99.99 & 99.99       \\
                             & proposed       & 99.95 & 99.95 & 99.96 & 99.96 & 99.96 & 99.96    \\
                             \hline
\multirow{4}{*}{VGG19}       & MagNet         & 58.20 & 48.52  & 57.73 & 48.24 & 57.84 & 48.45 \\
                             & DLA            & 88.51 & 88.77 & 83.96 & 84.62 & 83.57 & 84.18        \\
                             & PACA           & 99.98 & 99.98 & 99.97 & 99.97 & 99.99 & 99.99       \\
                             & proposed       & \textbf{100.00} & \textbf{100.00} & \textbf{100.00} & \textbf{100.00} & \textbf{100.00}    & \textbf{100.00}          \\
\bottomrule
\end{tabular}
\label{comparison}
\end{table*}

\subsection{Comparison with state-of-the-art}

In this section, we compare the performance of the proposed method in detecting AEs with the other three state-of-the-art (SOTA) and classic detection methods. MagNet \cite{meng2017magnet} learns the manifold of NEs and then reconstructs a new manifold. If there is an existing abnormal difference, the system will give a judge. It is suitable to compare our method with another reconstruct method. PACA \cite{chen2021adversarial} is a two-stream network construct that combines the deep feature of image and gradient from the victim classifier, these two data will be put into a designed binary classification network. DLA \cite{sperl2020dla} is a transfer learning which transforms the multi-classification task into a binary classification task, the binary classifier will learn the deep features of AEs from the victim classifier. All the methods are trained in the same condition including our method, the training data is generated by three known adversarial attack methods and the AEs in test data are also generated by the same attack methods. 

To test the generalization of different detection methods on different victim models, the test set is generated from the other two trained victim models, three different attack methods are in one test set. To improve the performance of the detection methods, we turn up the adversarial perturbation in the experiments.

The results of comparison experiments are shown in Table \ref{comparison} which contains detection accuracy and f1 Score. In every row of the table, the training data of the four methods is generated from one single victim model, in every column of the table, the test data is generated from other victim models, except that the MagNet is trained by the normal examples. We can conclude from table \ref{comparison} that the proposed method can outperform other methods in generalization to other victim classifiers except for the PACA. Different attack methods have different adversarial perturbations, the proposed method can extract the salient feature of the attack method, so we can detect the AEe from different perturbations, the proof experiment will be put in the next subsection.

\begin{figure*}[h]
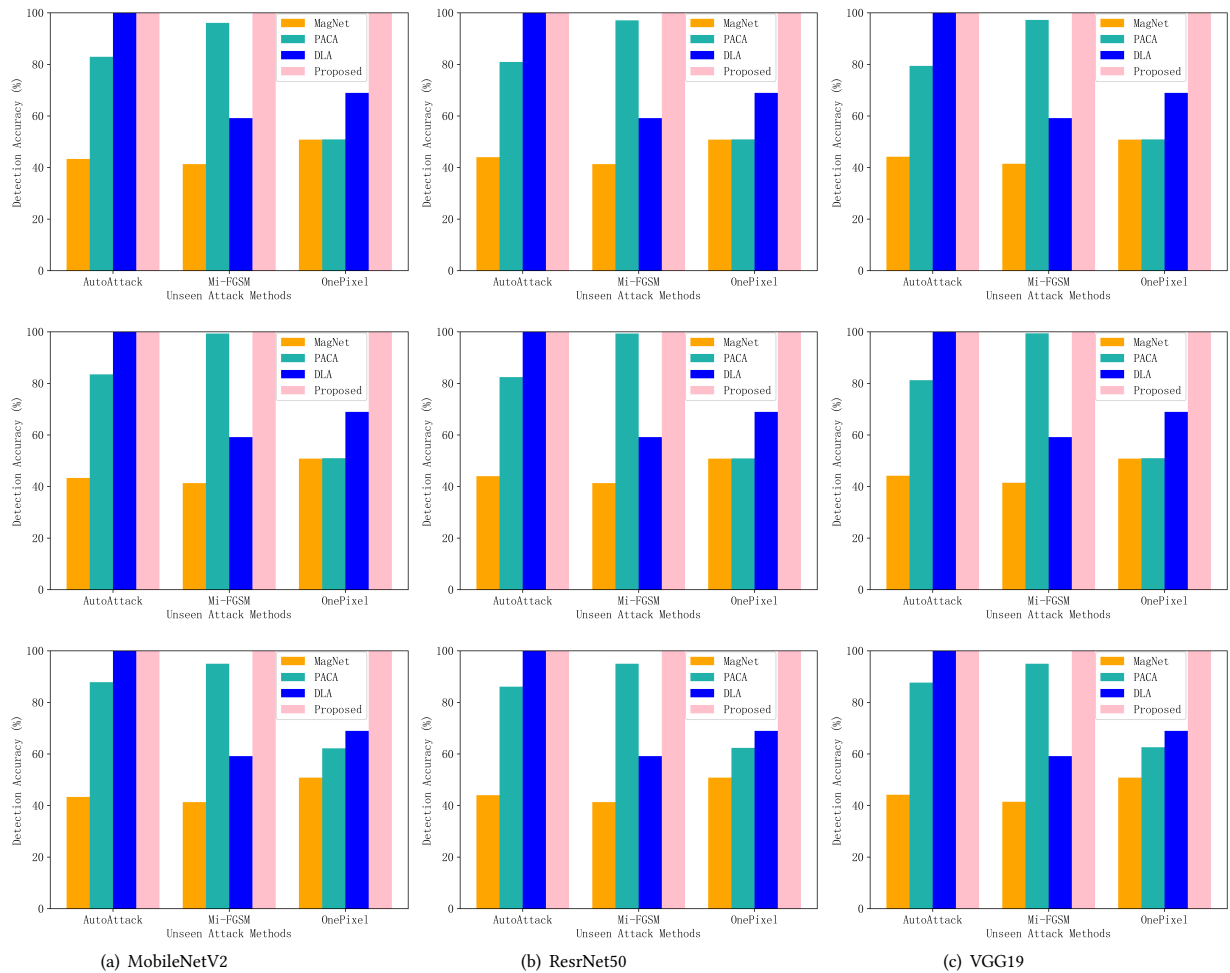

  \centering
  \subfigure[MobileNetV2]{
		\begin{minipage}[t]{0.25\linewidth}
			\centering
			\includegraphics[width=2.2in]{block1_1.pdf}\\
			\vspace{0.02cm}
			\includegraphics[width=2.2in]{block2_1.pdf}\\
			\vspace{0.02cm}
			\includegraphics[width=2.2in]{block3_1.pdf}\\
			\vspace{0.02cm}
		\end{minipage}%
	}
	\hspace{.3in}
	\subfigure[ResrNet50]{
	\begin{minipage}[t]{0.25\linewidth}
		\centering
		\includegraphics[width=2.2in]{block1_2.pdf}\\
		\vspace{0.02cm}
		\includegraphics[width=2.2in]{block2_2.pdf}\\
		\vspace{0.02cm}
		\includegraphics[width=2.2in]{block3_2.pdf}\\
		\vspace{0.02cm}
	\end{minipage}%
	}
	\hspace{.3in}
	\subfigure[VGG19]{
	\begin{minipage}[t]{0.25\linewidth}
		\centering
		\includegraphics[width=2.2in]{block1_3.pdf}\\
		\vspace{0.02cm}
		\includegraphics[width=2.2in]{block2_3.pdf}\\
		\vspace{0.02cm}
		\includegraphics[width=2.2in]{block3_3.pdf}\\
		\vspace{0.02cm}
	\end{minipage}%
	}
  \caption{Detection accuracy of three unseen adversarial examples. This experiment compares with other three SOTA detection methods.}
  \Description{A woman and a girl in white dresses sit in an open car.}
  \label{unseen}
\end{figure*}

We also consider testing the methods on other unseen adversarial attack methods. The attack method may not be easily known by us in reality, the generalization to other unseen attack methods should also be considered. So in this experiment, the trained methods above are to detect the newly generated AEs, and the result can be seen inFigure \ref{unseen}. AutoAttack \cite{2020Reliable}, MI-FGSM \cite{2022Boosting}, and OnePixel Attack \cite{2017One} are contained in the generated adversarial examples. In Figure \ref{unseen}, our method is better than others in detecting unseen adversarial attacks. These three methods can detect the large-perturbation unseen adversarial examples with not very high detection accuracy, but the detection accuracy of less-perturbation unseen adversarial examples is very low. In our method, both large-perturbation and less-perturbation can be detected with high accuracy. In discriminator of our method, the gradient-based perturbations and optimize-based perturbations are learned, so the MI-FGSM and AutoAttack can be detected, and the perturbation of the OnePixel Attack may have similarity with learned adversarial perturbations. Thus the unseen adversarial examples can be detected with high accuracy.

\subsection{AEs recovery}

To verify that the extracted adversarial perturbations by U-Net from adversarial examples are non-robust but predictive to the protected models, the AEs recovery experiments are set. In the victim models, the AE $x^{*}$ will be misclassified, $f(x^{*})\neq y$. But if the adversarial perturbations $\delta$ are subtracted from the adversarial examples, the results of the recovered examples should be corrected, $f(x^{*}-\delta)=y$. In this section, we recover AEs $x^{*}$ to be close the NEs $x$. The AE directly subtracts the extracted perturbation, we called it recovered example $x^{'}=x^{*}-\mathcal{G}(x^{*})$ , in this way the adversarial perturbation in the example is discarded. Under this hypothesis, the classification accuracy of the recovered examples should be close to the NEs.

In this experiment,  we test the classification accuracy of recovered examples from three victim models. The AEs are generated in the three attack methods described above. And the U-Net is trained by three different AEs generated from the victim model. The results are in Table \ref{recover}, we can see that the classification accuracy of AEs is low, but the recovered examples can be back to the high accuracy. It means that the proposed method can simulate the ground truth adversarial perturbation very well. The more detailed information of perturbation U-Net can extract, the more specific discriminator can learn.  If the inner information of adversarial perturbation can be extracted by the U-Net, the usage of the extracted perturbations can be expanded to many aspects, we will describe it in Section 4.5.
\begin{table}[htbp]
\caption{Performance of trained model on different datasets and models.}
\centering
\begin{tabular}{llccc}
\toprule
\multirow{2}{*}{DataSet} & \multirow{2}{*}{Model Name} & \multicolumn{3}{c}{Test Accuracy(\%)}\\ \cline{3-5}
          &         & original & adversarial & recovered \\ 
\midrule
\multirow{3}{*}{CIFAR-10}  & MovileNetV2  & 67.27 & 19.35 & 67.60  \\
                           & ResNet50     & 65.78 & 18.17 & 57.27  \\
                           & VGG19        & 79.47 & 19.42 & 63.68  \\
\multirow{3}{*}{ImageNet}  & MovileNetV2  & 66.73 & 5.69 & 53.78  \\
                           & ResNet50     & 62.47 & 5.80 & 52.62  \\ 
                           & VGG19        & 60.13 & 24.47 & 51.29  \\
\bottomrule
\end{tabular}
\label{recover}
\end{table}

\subsection{AEs calssification}

The experimental results above illustrate that the unknown source AEs can be detected by the proposed method with high accuracy. The following experiments will illustrate how model-free our method is. According to our hypothesis mentioned in Section 1, the adversarial perturbation of the same attack method is similar. So we prepare the experiment on AEs classification. We extract the adversarial perturbation from the input examples by trained U-Net and then train the discriminator to classify the extracted perturbations. In this experiment, we use PGD, C\&W, and DeepFool as the attack methods, and the U-Net was trained in Section4.3. 

\begin{figure}[h]
  \centering
  \includegraphics[width=1.1\linewidth]{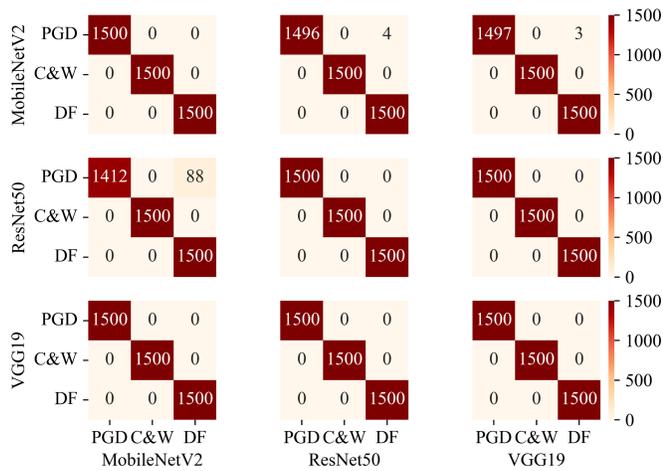}
  \caption{Confusion matrix of the adversarial examples classification.}
  \Description{A woman and a girl in white dresses sit in an open car.}
  \label{confuse}
\end{figure}

The extracted perturbation can be seen in Figure \ref{image_per}, three different attack methods have different perturbations, and we can see the obvious difference. The classification results are in Figure \ref{confuse}. The concrete classification categories are displayed in the confusion matrix. To test the generalization of the proposed method in classification, we also test the classification results in other test set generated from other victim models. In every row of the figure, the training data for method is generated from one victim model. And in every column of the figure, the test data is generated from other models. We can conclude from the confusion matrix that our hypothesis is right that the perturbations have the same texture in one attack method.

\begin{figure}[h]
  \centering
  \subfigure[NEs]{
		\begin{minipage}[t]{0.1\linewidth}
			\centering
			\includegraphics[width=0.7in]{1_ori.pdf}\\
			\vspace{0.02cm}
			\includegraphics[width=0.7in]{2_ori.pdf}\\
			\vspace{0.02cm}
			\includegraphics[width=0.7in]{3_ori.pdf}\\
			\vspace{0.02cm}
			\includegraphics[width=0.7in]{4_ori.pdf}\\
			\vspace{0.02cm}
			\includegraphics[width=0.7in]{5_ori.pdf}\\
			\vspace{0.02cm}
		\end{minipage}%
	}
	\hspace{.3in}
	\subfigure[PGD]{
	\begin{minipage}[t]{0.1\linewidth}
		\centering
		\includegraphics[width=0.7in]{1_pgd.pdf}\\
		\vspace{0.02cm}
		\includegraphics[width=0.7in]{2_pgd.pdf}\\
		\vspace{0.02cm}
		\includegraphics[width=0.7in]{3_pgd.pdf}\\
		\vspace{0.02cm}
		\includegraphics[width=0.7in]{4_pgd.pdf}\\
		\vspace{0.02cm}
		\includegraphics[width=0.7in]{5_pgd.pdf}\\
		\vspace{0.02cm}
	\end{minipage}%
	}
	\hspace{.3in}
	\subfigure[C\&W]{
	\begin{minipage}[t]{0.1\linewidth}
		\centering
		\includegraphics[width=0.7in]{1_cw.pdf}\\
		\vspace{0.02cm}
		\includegraphics[width=0.7in]{2_cw.pdf}\\
		\vspace{0.02cm}
		\includegraphics[width=0.7in]{3_cw.pdf}\\
		\vspace{0.02cm}
		\includegraphics[width=0.7in]{4_cw.pdf}\\
		\vspace{0.02cm}
		\includegraphics[width=0.7in]{5_cw.pdf}\\
		\vspace{0.02cm}
	\end{minipage}%
	}
	\hspace{.3in}
	\subfigure[DL]{
	\begin{minipage}[t]{0.1\linewidth}
		\centering
		\includegraphics[width=0.7in]{1_deepfool.pdf}\\
		\vspace{0.02cm}
		\includegraphics[width=0.7in]{2_deepfool.pdf}\\
		\vspace{0.02cm}
		\includegraphics[width=0.7in]{3_deepfool.pdf}\\
		\vspace{0.02cm}
		\includegraphics[width=0.7in]{4_deepfool.pdf}\\
		\vspace{0.02cm}
		\includegraphics[width=0.7in]{5_deepfool.pdf}\\
		\vspace{0.02cm}
	\end{minipage}%
	}
  
  \caption{Four columns are normal examples, extracted adversarial perturbation of PGD, extracted adversarial perturbations of C\&W, extracted adversarial perturbations of DeepFool, respectively. Images are chosen from ImageNet.}
  \label{image_per}
\end{figure}

\subsection{Robustness analysis}

In the real-world scene, the input examples are easy to be distorted. Therefore, if the NEs are polluted by salt and pepper noise which is similar to adversarial perturbation visually, the accuracy of detection may degenerate. In this section, to test the robustness of our detection methods, we add pepper and salt noise to NEs to simulate the distortion of images.

\begin{figure}[h]
  \centering
  \includegraphics[width=1.1\linewidth]{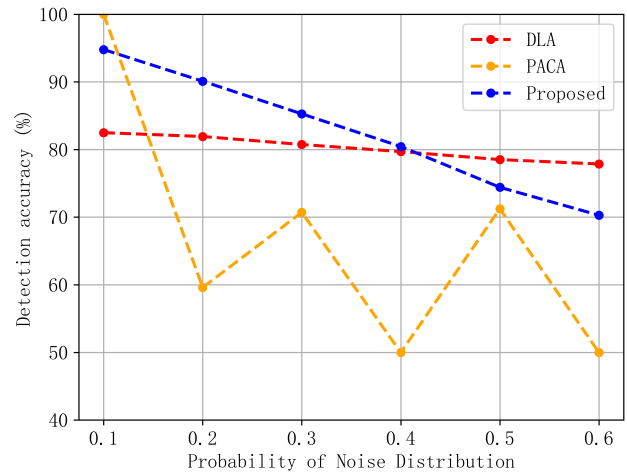}
  \caption{Detection accuracy on different noise density added to the normal examples.}
  \label{robust}
\end{figure}

In this experiment, the adversarial examples are generated from VGG19 and the images are chosen from ImageNet. The NEs on test set are added with different densities of pepper and salt noise. The higher the density is added, the more seriously the NE is distorted. The value of the density ranges from 0.1 to 0.6. If the density is greater than 0.6, the image will be abandoned for it is seriously damaged. Compared with the well-performed DLA and PACA, the result is shown in Figure \ref{robust}. According to the line chart, three methods are all influenced by the noise, among which PACA is more sensitive to the noise and its detection accuracy is unstable. The accuracy of DLA and our method reduces a little, but our method keeps higher accuracy than that of DLA. Consequently, our method may be affected by the random noise, but our method is still robust while detecting the examples polluted by the slight noise.

\section{Conclusion}
In this paper, a model-free adversarial perturbation extractor and an adversarial example detector are proposed, both of which are free from querying the victim models. We concentrate on extracting the high-dimension and adversarial features of the input examples by U-Net, and detect the adversarial examples using their corresponding extracted high-dimension features. The adversarial perturbations projected by non-robust but predictive features of the victim classifiers are extracted as much as possible by our method. With perturbation extracted, we can restore the adversarial examples, thus the AEs can be recycled and the quality of the recovered examples is close to the normal examples. The difference in perturbation between AEs and NEs is more obvious than that among the NEs, and the detection accuracy in large-perturbation attacks can reach 100\%. We also test the generalization of our method on unseen methods which include the few-perturbation attack, and experimental results show superior performance over other SOTA methods. To verify the independence of our method on victim models, the validation experiment is set that the detector is used to classify the extracted perturbations generated from three different models and the results met expectations. The perturbation generated by same attack method is similar, even though the examples are generated from different models. Besides, the difference of the different attacks is fully exploited by detectors to distinguish the categories of the adversarial attacks. Considering the acceptable pollution to the NEs after extraction which does not affect the performance, the robustness of our method is also guaranteed. However, there are still some problems in AEs recovery, for the test accuracy of the recovered examples is not good enough to achieve the same accuracy as the original samples. In future work, we will work on improving the quality of the recovered adversarial examples based on this work.


\begin{acks}
This work was supported by the National Natural Science Foundation of China (Grant No. 61300055), Zhejiang Natural Science Foundation (Grant No. LY20F020010), Ningbo Natural Science Foundation (Grant No. 202003N4089).
\end{acks}

\printbibliography

\end{document}